# PathReasoning: A multimodal reasoning agent for query-based ROI navigation on whole-slide images

Kunpeng Zhang[1,*], Hanwen Xu[1,#,*], Sheng Wang[#]

[1]School of Computer Science and Engineering, University of Washington, Seattle, WA

*Equal contribution

[#]Email (Corresponding authors): xuhw@cs.washington.edu, swang@cs.washington.edu

## Summary

Deciphering tumor microenvironment from Whole Slide Images (WSIs) is intriguing as it is key to cancer diagnosis, prognosis and treatment response. While these gigapixel images on one hand offer a comprehensive portrait of cancer, on the other hand, the extremely large size, as much as more than 10 billion pixels, make it challenging and time-consuming to navigate to corresponding regions to support diverse clinical inspection. Inspired by pathologists who conducted navigation on WSIs with a combination of sampling, reasoning and self-reflection, we proposed "PathReasoning", a multi-modal reasoning agent that iteratively navigates across WSIs through multiple rounds of reasoning and refinements. Specifically, starting with randomly sampled candidate regions, PathReasoning reviews current selections with self-reflection, reasoning over the correspondence between visual observations and clinical questions, and concludes by proposing new regions to explore. Across rounds, PathReasoning builds a reasoning chain that gradually directs attention to diagnostically relevant areas. PathReasoning turns each whole slide into a sequence of question-guided views, allowing the model to efficiently find informative ROIs within a fixed number of steps, without the need for dense pixel-level annotations. PathReasoning can substantially outperform strong ROI-selection approaches by 6.7% and 3.1% of AUROC on subtyping and longitudinal analysis tasks. The high-quality ROIs further support accurate report generation on breast cancer, significantly outperforming the standard GPT-4o by 10% in accuracy. PathReasoning prioritizes question-specific regions and constructs interpretable reasoning chains, supporting efficient slide review, consistent diagnostic interpretations, comprehensive reporting, and evidence traceability in digital pathology.

**Introduction**

The Whole Slide Images (WSIs) for pathology serves as the gold standard for cancer diagnosis[1–6]. Deciphering the tumor microenvironment from WSIs is of paramount importance, informing the current cancer stage, potential treatment therapy and future treatment response[7–9]. These gigapixel images are often extremely large, as much as more than 10 billion pixels[10], making it difficult to search for the corresponding regions which support clinical decision making[11]. In reality, its extreme size prevents pathologists from investigating all details simultaneously, resulting in a long analysis process[12–14]. In fact, pathologists have to navigate through the WSI by sampling and carefully examining several specific regions of interest (ROIs) for multiple rounds. This process is extremely long and tedious, taking 4-15 minutes[15] on average for analyzing a single slide, suggesting the pressing need for automatic AI-powered navigation.

However, most existing computational methods[16–18] for analyzing WSIs lack the built-in reasoning capability for navigation. First, they rely on either the randomly sampled small patches or patches selected by a heuristic-based approach, without looking through the entire slide, making it easier to miss critical regions. These approaches often include irrelevant areas and may miss important features[19]. Second, most existing approaches exploit a singe-turn navigation approach, without iteratively refining selected regions through thinking and reflection, often resulting in selecting sub-optimal regions. Therefore, these limitations suggest the growth area of new methods with the ability to explore the entire slide through multi-turn refinements, echoing standard practice where experienced pathologists examine WSIs with complex internal reasoning.

As a result, we propose PathReasoning, a question-conditioned, iterative ROI-selection agent powered by GPT-4o that simulates expert reasoning to navigate gigapixel WSIs[20] **(Fig. 1)**. PathReasoning extends single-turn ROI selection into a multi-turn and multi-scale navigation process on WSIs. The key idea of PathReasoning is to create multiple, consecutive views of a slide by iteratively proposing new regions at a different position and magnification level[21], with each view forming part of a consistent evidence chain used for downstream reasoning, thereby effectively augmenting informative evidence while filtering irrelevant regions. We design special prompts to expose the multi-modal reasoning[22] capability of frontier models for WSI navigation. Our model employs an iterative think-act-reflect loop to simulate the internal reasoning process of human experts[23]. Specifically, for each clinical question, our approach consists of three steps: (1) PathReasoning adaptively infers ROIs by associating the visual observations to clinical knowledge, (2) PathReasoning automatically retrieves corresponding regions by specifying the coordinates and magnification level of regions, (3) PathReasoning refine the chosen region with self-reflection on whether it contains sufficient information for clinical question answering. Unlike heuristic selection or uniform patch extraction, PathReasoning actively reasons about where to look next, enabling more targeted and informative region selection without relying on dense annotations.

We systematically evaluated PathReasoning on both standard cancer classification tasks and clinical report generation. On classification tasks across TCGA[24] cohorts, such as cancer subtyping, we observed significant improvements with its built-in reasoning capability compared to competing approaches in both accuracy and $F_1$ score. We further confirm this by evaluating the same ROIs with multiple foundation models. Using light-weight classifiers (k-NN[25] and logistic regression[26] heads), we found that PathReasoning outperformed comparison approaches, with five points absolute AUROC improvements on average across three commonly used foundation models. Our experiments demonstrated improved performance as more reasoning steps are deployed, echoing the recent progress of LLMs on test-time scaling[27]. Furthermore, we showed that PathReasoning can stratify patients into risk groups with significantly different survival distributions, achieving consistent Kaplan–Meier[28] separation across TCGA cohorts (log rank $p$-value $< 0.05$). Having established improvements on classification, we next evaluate generative objectives that aggregate visual evidence into text. On free-text pathology report generation, PathReasoning produces more faithful and clinically complete reports while using the same amount of visual input as baseline methods. We then move to a clinically grounded evaluation with pathology checklist[29], where PathReasoning yields higher item-level accuracy and fewer omissions. Finally, in WSI-VQA[30], a breast cancer (BRCA) LLM-generated visual question answering (VQA) dataset from TCGA cohorts, PathReasoning outperformed existing methods, with improvements of 9.5% and 9.2% in Proliferation and Lymph Nodes categories. Collectively, PathReasoning speeds up the navigation process across WSIs, showing substantial improvements against existing approaches on cancer subtyping, clinical report generation, and survival risk predictions, paving the way to democratize precision medicine to underserved communities where experienced pathologists are scarce.

## Results
### Overview of PathReasoning

PathReasoning takes a WSI together with a textual clinical query (e.g., subtype classification, survival risk prediction, or pathology report generation) as input and outputs one or more ROIs through an iterative reasoning process **(Fig. 1a)**. The navigation process is organized into a Think–Act–Reflect loop[31], where the model hypothesizes candidate regions based on global context and prior ROI history (Think), selects and examines new coordinates on the WSI (Act), and evaluates whether the chosen ROI provides sufficient information to stop or continue exploring (Reflect). For example, PathReasoning first zooms into a region and notices scattered cells that lack the signs of suspected abnormality. As a result, it may zoom out to re-examine the broader context and then move to a new region, which displays more consistent structural patterns that support a clearer diagnosis.[32] This design allows PathReasoning to simulate the workflow of human pathologists by proposing candidate regions, revising earlier decisions, and progressively narrowing down the most informative areas.

Beyond its core design, PathReasoning demonstrates strong generality across tasks. The iterative ROI navigation mechanism can be flexibly applied to discriminative tasks such as cancer subtyping and survival risk prediction, as well as generative tasks including visual question answering (VQA)[33] and clinical report generation. In addition, the agent supports different output modes for these tasks, returning either a single final ROI or multiple intermediate ROIs, which provides flexibility for both discriminative and generative applications without requiring task-specific re-training.

To evaluate our results, we compared PathReasoning with two comparison approaches: a majority-vote[34] approach that aggregates predictions from randomly sampled ROIs **(Supplementary Fig. 4a)** and a single-turn GPT-based selection approach **(Supplementary Fig. 4b)**. These approaches provide simple but reasonable reference points for ROI selection. Across TCGA[24] cohorts, we observed that both approaches consistently underperformed PathReasoning. This highlights the advantage of iterative reasoning in navigating heterogeneous WSIs and motivates our large-scale evaluation across multiple cancer types and downstream tasks.

**PathReasoning improves cancer subtyping accuracy across most TCGA cohorts**
We first evaluated PathReasoning on cancer subtyping using 13 TCGA cohorts across both common and rare cancer types **(Fig. 2a)**. We compared three approaches: a majority-vote approach, a single-turn GPT approach, and the full PathReasoning framework. The majority-vote approach aggregates predictions from multiple ROIs that are randomly sampled across the tissue slide; each ROI is classified independently and the final subtype prediction is obtained by voting across all sampled regions. The single-turn GPT approach instead prompts the model to directly select one ROI based on a global thumbnail of the slide, and the downstream prediction is made solely from this single selection. In contrast, PathReasoning iteratively explores slides through the Think–Act–Reflect loop, refining ROIs step by step and terminating early when sufficient evidence is reached, with a cap of 10 iterations.

Overall, PathReasoning substantially outperformed both baselines across 10 out of 13 cancer types, achieving on average a 17.9% accuracy improvement over the single-turn GPT approach, and likewise surpassing the comparison approaches in macro $F_1$ score for 10 out of 13 cancer types **(Fig. 2c, Supplementary Fig. 1a)**. Representative case studies further highlight these improvements **(Fig. 2d)**. In these cases, PathReasoning correctly identified informative ROIs that enabled accurate cancer subtype classification, whereas both the single-turn GPT approach and majority-vote approach misclassified the same slides due to selecting less discriminative regions. Performance improvements were especially evident for multi-class or morphologically heterogeneous cohorts such as RCC[35] (CCRCC vs. CHRCC vs. PRCC) and ESO (ESCA vs. ESCC vs. STAD), suggesting that iterative ROI navigation is particularly effective when subtyping relies on interpreting localized structural patterns that are not captured by a single region alone. Despite substantial biological and morphological heterogeneity, PathReasoning consistently improved both metrics across nearly all cohorts, suggesting that the model has strong generalization ability for most tumor types. We then investigated the effect of iterative reasoning steps on classification performance **(Fig. 2b)**. Accuracy improved steadily as the number of reasoning steps increased, with reduced improvements after around ten iterations. By revisiting earlier decisions and progressively focusing on more informative regions, the agent overcomes the limitations of single-turn ROI selection.

We next verified that the ROIs identified by PathReasoning capture informative diagnostic value rather than reflecting random visual patterns **(Fig. 2f, Supplementary Fig. 1d)**. Using the pretrained foundation models Prov-GigaPath[19], UNI[36], and H-optimus-0, we compared ROI embeddings from PathReasoning against those from the single-turn GPT approach with logistic regression[26] and k-nearest neighbors[25] (k-NN) using light-weight classifiers. Across both classifiers and all three encoders, PathReasoning achieved stronger performance, with improvements in 10 of 13 cancer types under logistic regression and k-NN. Averaged across cancer types using k-NN classifier, the mean AUROC improvement of PathReasoning over the single-turn GPT approach was 0.03 for Prov-GigaPath, 0.07 for UNI, and 0.05 for H-optimus-0. Improvements were most pronounced in GLIOMA (GBM vs. ODG) and CERVIX (CESC vs. ECAD), where subtype features are highly localized. These results demonstrate that the agent's iterative ROI selection not only drives end-to-end improvements but also produces compact and discriminative embeddings that remain effective when transferred to independent models.

**PathReasoning enriches clinical report generation in both narrative quality and factual coverage**

To further illustrate the generalizability and clinical relevance of PathReasoning, we extended our evaluation from cancer subtyping to clinical report generation **(Fig. 3a)**. This task involves generating full clinical pathology reports from WSIs, requiring the agent to describe relevant morphological findings and integrate them into consistent diagnostic summaries. We first evaluated report generation using GPT-based evaluation scores, where the large language model (GPT-4o)[37] was asked to compare each ROI-generated report with its ground-truth report[38] and assign a score from 0 to 10, with higher scores indicating greater semantic similarity between the two reports. The scoring reflects linguistic fluency, coherence, and the consistency of diagnostic content between the generated and reference narratives. PathReasoning consistently achieved higher GPT evaluation scores than both comparison approaches across 13 cancer types, with average improvements of roughly 0.4 - 0.6 points on the 10-point scale **(Fig. 3e)**. We observed the improvements tend to be larger on tasks where pathological features are concentrated in local, high-salience ROIs that benefit from targeted selection, such as GLIOMA (GBM vs. ODG) and HEP (CHOL vs. HCC). These results highlight the strength of PathReasoning in supplying contextually rich evidence for clinical text generation, and further suggest that its iterative ROI navigation is particularly advantageous in cancers where diagnostic signals are localized to small but highly informative regions.

Beyond overall report quality, we also wanted to know whether the generated clinical reports captured the important clinical features that pathologists document during diagnosis. For each cancer type, we conducted a checklist-based evaluation using several selected questions from its corresponding TCGA enrollment forms, which record key attributes such as tumor architecture, mitotic activity, and presence of necrosis. By comparing ROI-generated reports and ground-truth reports against these checklists, we directly tested the extent to which PathReasoning preserves clinically salient details essential for diagnostic interpretation. PathReasoning again consistently outperformed both comparison approaches across most cancer types, improving factual accuracy by several percentage points on average **(Fig. 3d)**. This is particularly evident in cohorts such as COLON (COAD vs. READ) and BRCA (IDC vs. ILC), where many checklist attributes, including tumor architecture, margin status, and glandular patterns, can be directly verified from localized regions of the slide. Therefore, these results indicate that the checklist evaluation confirms PathReasoning's ability to preserve clinically relevant details in generated reports, which proves our framework can deliver a clear advance in pathology text generation by combining accurate factual grounding with coherent diagnostic narratives.

**PathReasoning enhances fine-grained text–vision reasoning in WSI-VQA**
After validating PathReasoning on TCGA pathology reports and clinical checklists, we next evaluated it on the WSI-VQA[30] benchmark for breast cancer (BRCA). Unlike real pathology reports or enrollment checklists, the WSI-VQA dataset is constructed from TCGA slides but reformulated into structured question and answer pairs by large language models and expert editing. This format enables systematic benchmarking across categories of questions, ranging from morphological features to clinical attributes and outcomes. Across all BRCA questions, PathReasoning achieved higher accuracy than both comparison approaches, with an average improvement of about 10% compared with the single-turn GPT approach **(Fig. 3c)**. We next investigated the improvements across several clinical categories. We found that the largest improvements were in diagnosis, grading, and size/multiplicity questions, which require detailed interpretation of local morphology features. For example, in a BRCA case study **(Fig. 3b)**, the agent shifted its focus from stromal areas to tumor-rich ROIs displaying marked nuclear atypia and organized glandular structures, leading to answers consistent with the ground-truth diagnosis and histologic grade. On the other hand, performance improvements were more modest in categories such as stage and biomarker-related questions, where relevant information is distributed more broadly or not directly observable in H&E slides. Overall, PathReasoning consistently delivered robust advantages across the benchmark, demonstrating its ability to adapt step-wise spatial reasoning to the structured question–answer format of WSI-VQA and highlighting its generalizability for diverse pathology reasoning tasks.

**PathReasoning identifies prognostic regions for survival risk prediction**

To further demonstrate the clinical relevance of PathReasoning, we evaluated its ability to predict patient survival from WSIs by assigning each case to one of three risk groups: low risk (more than 36 months), medium risk (12–36 months), or high risk (less than 12 months) **(Fig. 4a)**. Predictions are made by examining the ROI or set of selected ROIs and using them as the basis for risk stratification[39], thereby testing whether the chosen regions contain prognostically informative morphology with potential value for treatment planning. We evaluated this task using Kaplan–Meier (KM)[28] survival analysis, plotting KM curves and evaluating separation with log-rank tests and hazard ratios. We focused this analysis on BRCA (IDC vs. ILC), LUNG (LUAD vs. LUSC), and COLON (COAD vs. READ), as these cancer types in TCGA provide sufficiently large cohorts with long-term follow-up data. In the BRCA cohort, PathReasoning showed clear survival separation, with statistically significant differences in both IDC and ILC subtypes (log-rank $p < 0.05$) **(Fig. 4c)**. By contrast, neither the single-turn GPT approach nor the majority-vote approach achieved significant separation **(Fig. 4d–e)**. Similar trends were observed in COLON (COAD versus READ) and LUNG (LUAD versus LUSC), where PathReasoning consistently produced stronger curve separation than both approaches **(Fig. 4f–k)**. For example, in a COLON case study **(Supplementary Fig. 3a-c)**, the agent initially focused on non-diagnostic stroma but progressively shifted toward tumor-rich regions. These regions displayed glandular organization and nuclear atypia, which are key features linked to clinical outcomes. Eventually, the agent stabilized on ROIs that matched the ground-truth features. These results confirm that PathReasoning's selected ROIs can capture prognostically informative morphology, enabling robust survival stratification across multiple cancer types and providing promising spatial reasoning for precision oncology.

**Discussion**

We have proposed PathReasoning, an agentic framework for query-driven region selection on gigapixel WSIs. PathReasoning iteratively navigates a slide with a think–act–reflect loop to surface one or more ROIs that are most informative for the clinical query. We have shown strong performance on cancer subtyping, clinical report generation, and survival risk prediction, consistently outperforming majority-vote and single-turn GPT selection approaches across multiple TCGA cohorts and with different foundation models (e.g., Prov-GigaPath, UNI, and H-optimus-0). Collectively, PathReasoning turns gigapixel slides into concise, human-auditable evidence for downstream models, providing a general, training-light system that can be paired with existing vision–language systems to accelerate reliable computational pathology.

**Limitations of the Study.** PathReasoning has at least two limitations that we plan to address in future work. First, the number of ROIs is not universally optimal across tasks or slides. In the current implementation we return a small, fixed set of ROIs (e.g., 1–5). While this setting works well for many classification settings, it may undersample diffuse or multi-focal signals and is particularly restrictive for generation tasks (e.g., report writing) that benefit from broader evidence. Besides, the top-k ROIs found early in the trajectory are not guaranteed to be globally optimal. Therefore, we will replace the fixed-step termination with a task-aware adaptive stopping strategy that weighs marginal information improvement, feature diversity, and compute cost, and we will explore aggregation schemes that explicitly encourage complementary ROIs (e.g., coverage constraints across tissue compartments and multi-scale fusion). Second, coordinate selection is heuristic[40] and offers limited guarantees on optimality or transparency. ROI coordinates are currently produced by prompt-driven rules, which can revisit redundant regions or miss atypical yet informative areas. We will formalize selection as a constrained sequential decision problem, with rewards that proxy downstream utility (e.g., validation improvement or estimated information gain), diversity, and latency, and we will learn policies via offline or budgeted reinforcement learning[41] and contextual bandits using the trajectories we already collect.

There are a few existing ROI selection strategies for WSIs, such as saliency or activation heatmaps followed by Top-K[42] picking with non-maximum suppression, and text-retrieval–style ranking of patches using vision–language encoders and a query.[43] PathReasoning differs in at least two important ways. First, most existing methods are single-shot and local. In contrast, PathReasoning performs iterative selection with memory and reflection, which helps it navigate across scales and cover multi-focal or diffuse signals that a one-turn Top-K can miss. Second, retrieval approaches are sensitive to prompt phrasing and vocabulary; PathReasoning accumulates intermediate evidence and adapts its trajectory when patterns are sparsely expressed, which improves recall on long-tail cases. Compared with heatmap Top K selection method, PathReasoning chooses regions sequentially in a think–act–reflect loop and updates its context after each iteration. It optimizes a tradeoff between relevance and diversity, so it avoids picking many near-duplicate tiles that simply follow the hottest spots on a heatmap. It adapts the number of ROIs to the marginal information gain instead of using a fixed K, which helps on slides with diffuse or multifocal signals. It also explores across scales and returns a clear selection trace with short explanations, making the decision path easier to inspect. In heterogeneous slides and under tight ROI budgets, this adaptive procedure is more sample-efficient than a single turn over the top heatmap scores.

# Figure Legends

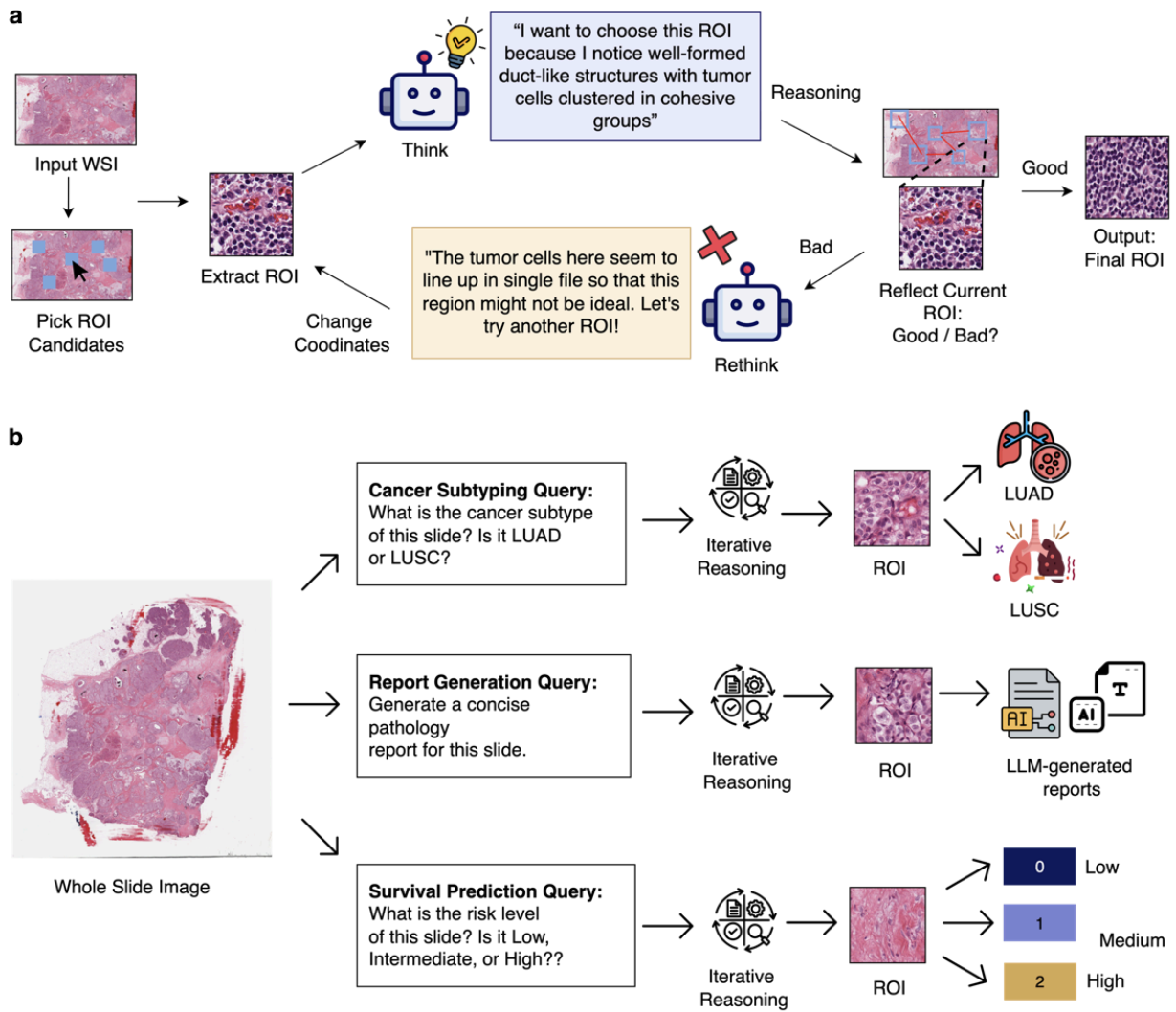

**Fig. 1 Overview of the PathReasoning agent and tasks. a,** Schematic of PathReasoning's iterative Think–Act–Reflect process for ROI selection on whole-slide images (WSIs). The reasoning cycle continues until a final ROI is identified, which is then used for downstream tasks. **b,** Schematic of downstream tasks evaluated by PathReasoning, including cancer subtyping, report generation, and survival risk prediction.

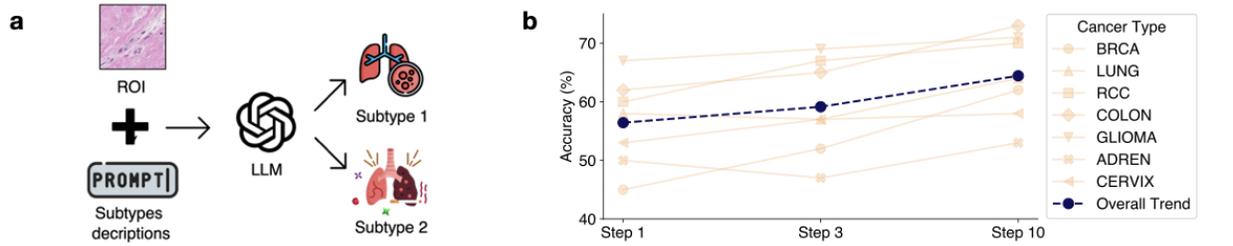

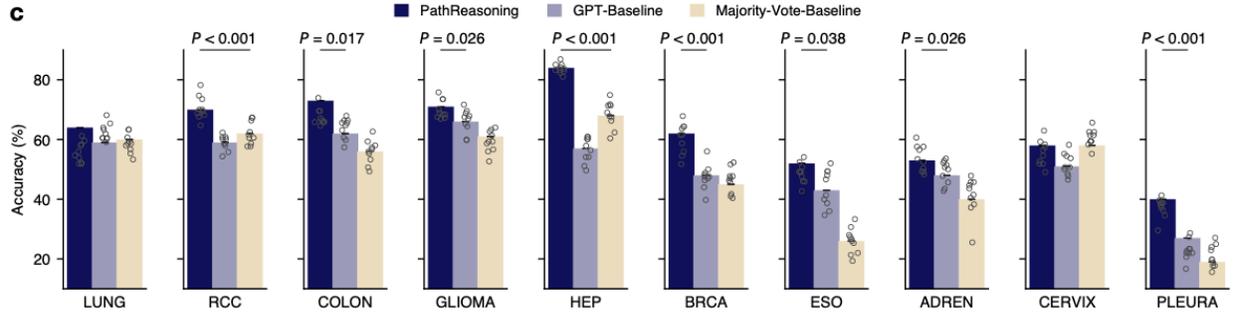

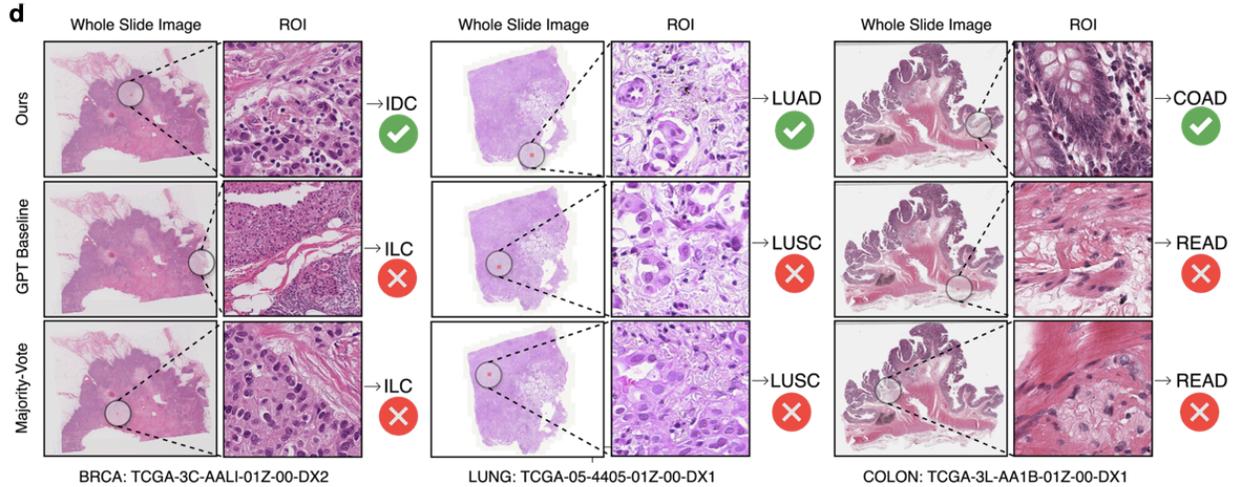

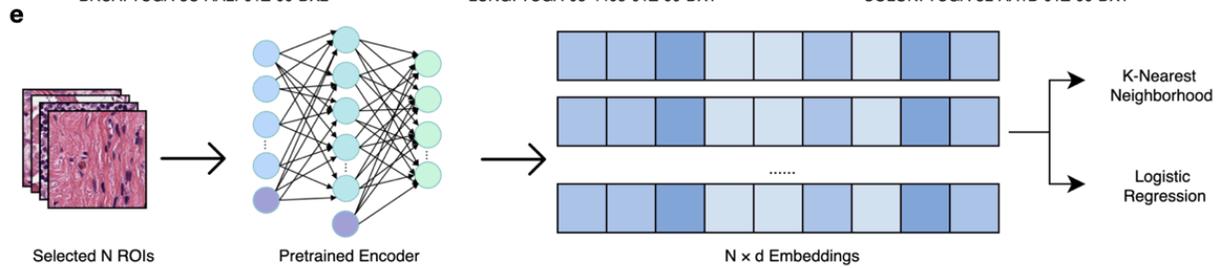

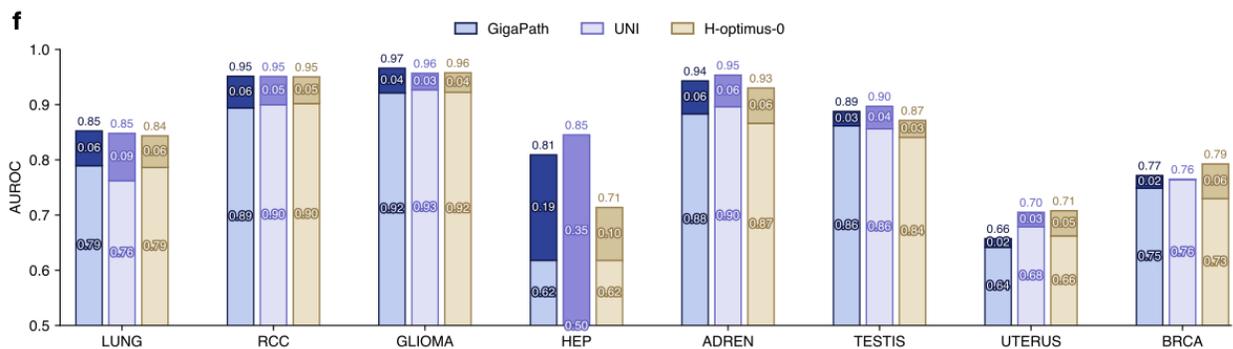

**Fig. 2 Cancer subtyping performance across TCGA cohorts. a,** Schematic illustration of the subtyping framework: ROIs are selected and paired with subtype description prompts, and then processed by a large language model to predict cancer subtype. **b,** Bar plots showing accuracy of PathReasoning, single-turn GPT approach, and majority-vote approach across 10 TCGA cancer types. Error bars denote standard deviation, gray dots represent bootstrap samples, and *p*-values are from paired t-tests. **c,** Bar plots showing macro F1 score across the same cancer types, with the same notation for error bars, bootstrap dots, and *p*-values. **d,** Line plot showing accuracy of PathReasoning as a function of the number of reasoning iterations (1, 3, 10). Each light line corresponds to a cancer type, and the dark dashed line indicates the overall trend. **e,** Schematic illustration of the foundation model evaluation task. Selected ROIs are preprocessed and passed through the pretrained Prov-GigaPath encoder to obtain 1536-d embeddings, which are then evaluated using lightweight classifiers (logistic regression or k-NN). **f,** Stacked bar plots showing AUROC of k-nearest-neighbors classifiers trained on ROI embeddings from three pretrained encoders (Prov-GigaPath, UNI, H-optimus-0) across 7 TCGA cancer types.

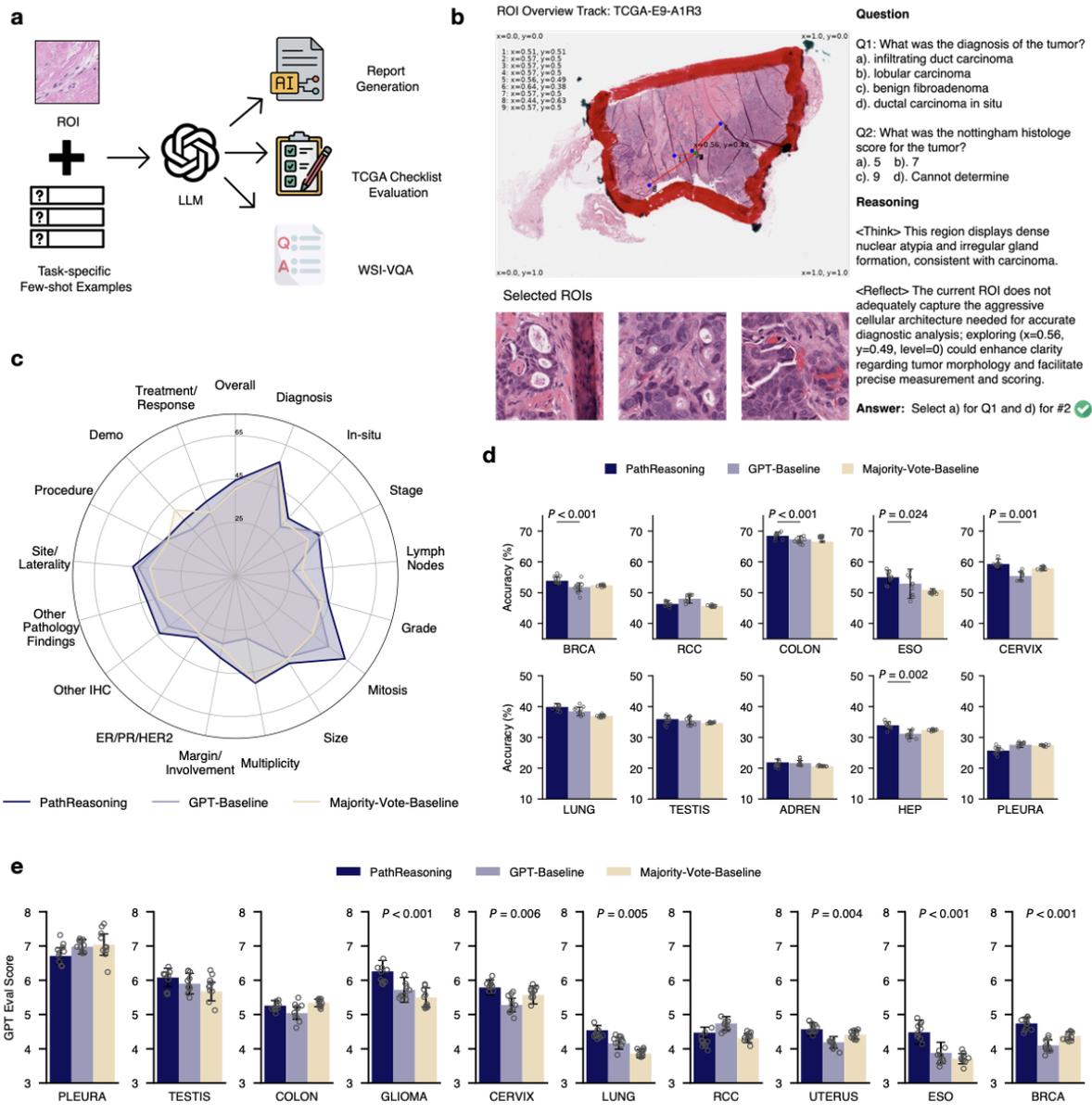

**Fig. 3 Evaluation of pathology report generation and visual question answering. a,** Schematic illustration of three evaluation tasks: GPT-eval Score evaluation for report generation, WSI-VQA evaluation, and TCGA checklist evaluation. **b,** Case study example from BRCA WSI-VQA showing the selected ROI regions, high-magnification patches, the question–answering process of the agent, and final predicted answers compared with ground-truth labels. **c,** Radar plot showing accuracy of WSI-VQA grouped by clinical and pathological question categories, with "Overall" reflecting aggregate accuracy. Shaded regions represent mean accuracy for each method. **d,** Bar plots showing checklist-based evaluation accuracy of generated reports across 10 cancer types, with the same notation for error bars, bootstrap dots, and $p$-values. **e,** Bar plots

showing GPT-based evaluation scores of generated pathology reports across 10 TCGA cancer types. Error bars denote standard deviation, gray dots represent bootstrap samples, and *p*-values are from paired *t*-tests.

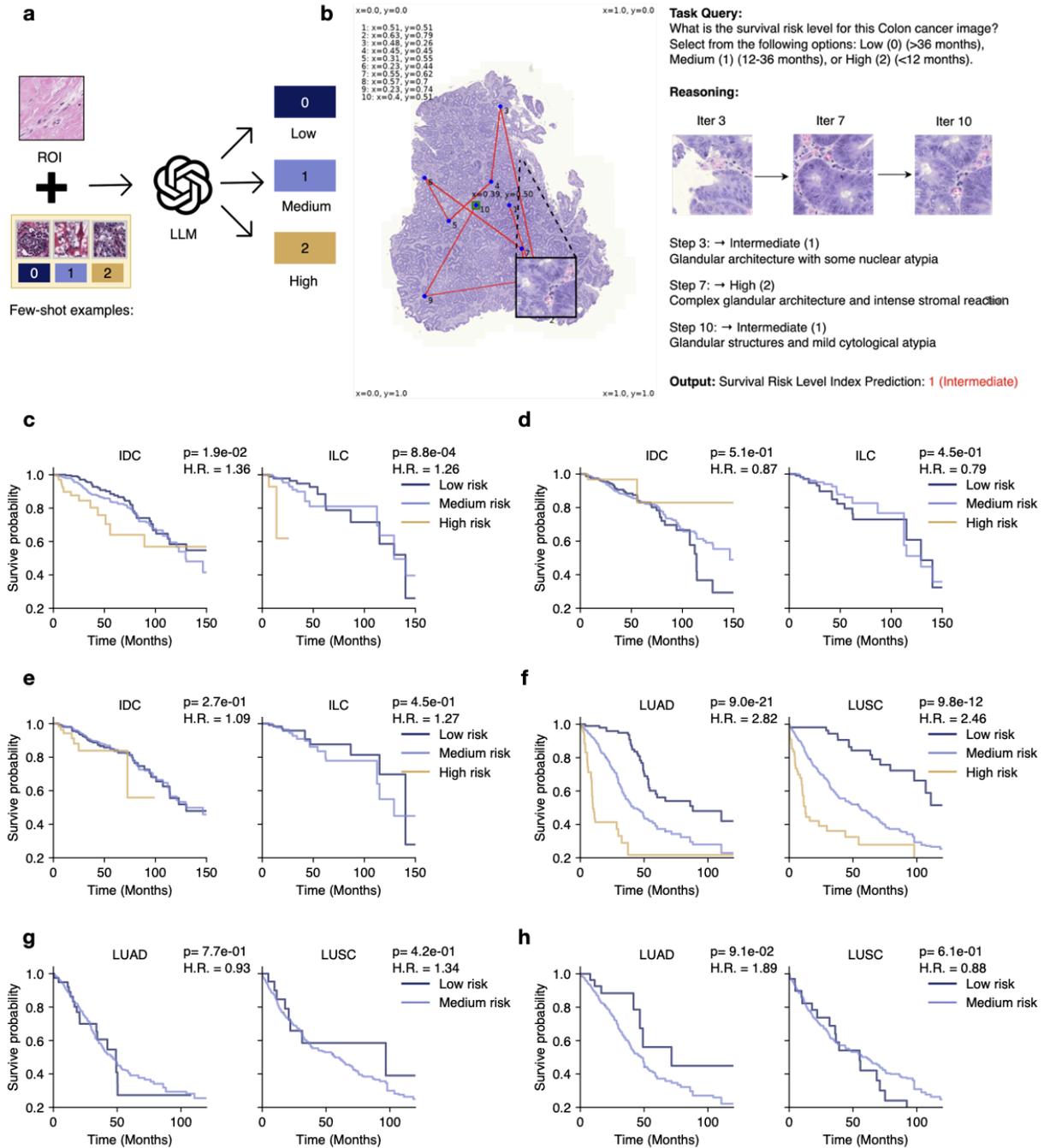

**Fig. 4 Survival risk stratification with PathReasoning. a,** Schematic illustration survival-risk prediction: ROIs with few-shot examples (three per risk level: 0/1/2 for Low/Medium/High) are provided to an LLM to infer patient risk. **b,** Case study on a colon cancer slide illustrating iterative ROI search: the agent proposes and refines candidate regions across iterations (examples from iters 3, 7, and 10 shown with morphological rationale) and outputs the final risk level. **c–e,** Breast carcinoma (BRCA) cohorts Kaplan–Meier curves comparing **c,** PathReasoning, **d,** single-turn GPT approach, and **e,** majority-vote approach. Curves are stratified into Low, Medium, and High risk. **f–h,** Lung cancer (LUNG) cohorts Kaplan–Meier curves for **f,** PathReasoning, **g,** single-turn GPT approach, and **h,** majority-vote approach, with the same risk stratification. *P*-values are from log-rank tests and hazard ratios (H.R.) from Cox models. Time is shown in months.

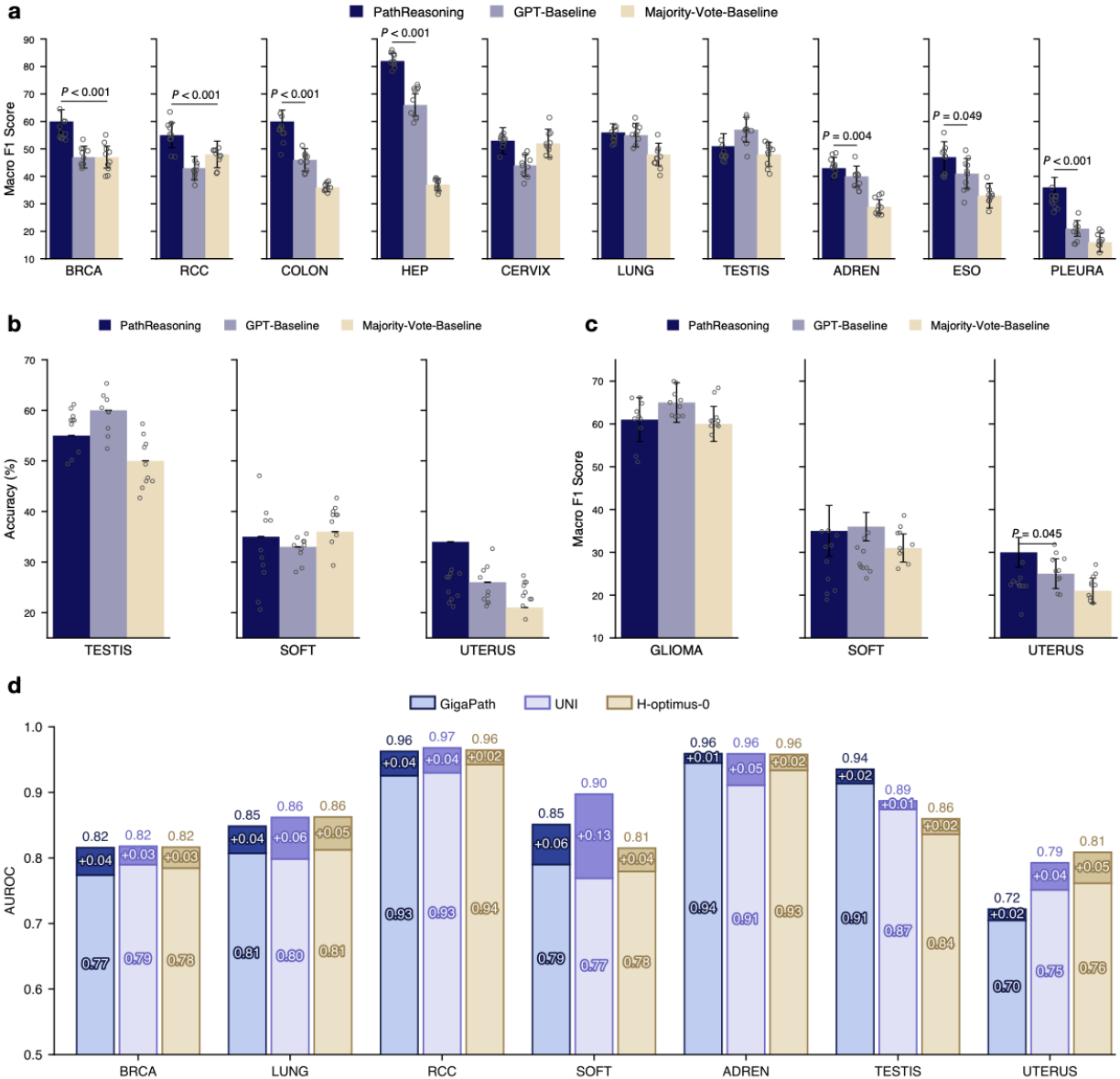

**Supplementary Figure 1. Performance on cancer subtyping and classification across additional cohorts. a,** Case studies highlighting representative examples where PathReasoning selects ROIs leading to correct subtype predictions, while single-turn GPT approach and majority-vote approach focus on misleading ROIs resulting in misclassification. **b,** Bar plots showing accuracy of PathReasoning, single-turn GPT approach, and majority-vote approach TESTIS, SOFT, and UTERUS. Error bars denote standard deviation, gray dots represent bootstrap samples, and p-values are from paired t-tests. **c,** Bar plots showing macro F1 scores on GLIOMA, SOFT, and UTERUS cohorts with the same notation for error bars, bootstrap dots, and p-values. **d,** Stacked bar plots showing AUROC of logistic regression classifiers trained on ROI embeddings from three pretrained encoders (Prov-GigaPath, UNI, H-optimus-0) across 7 TCGA cancer types.

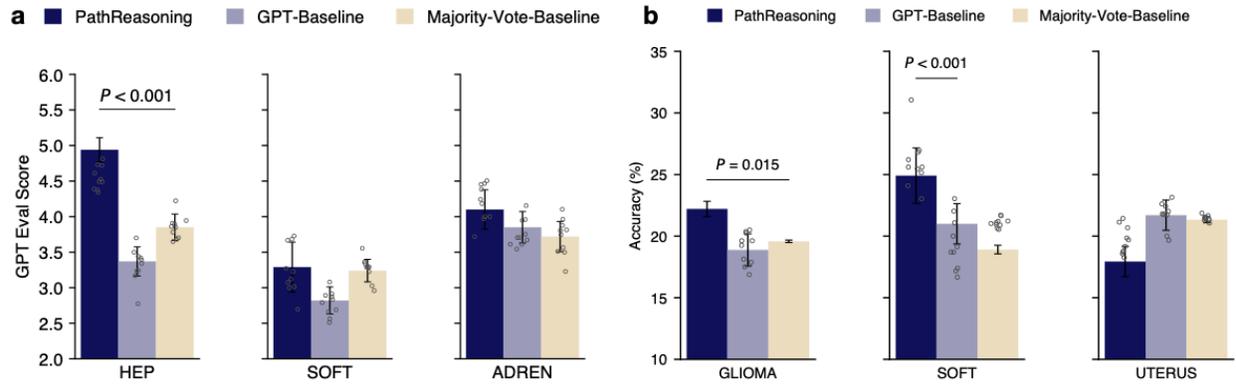

**Supplementary Figure 2. Supplementary results for report evaluation tasks. a,** Bar plots showing GPT-based evaluation scores of generated pathology reports across HEP, SOFT, ADREN. Error bars denote standard deviation, gray dots represent bootstrap samples, and p-values are from paired t-tests. **b,** Bar plots showing checklist-based evaluation accuracy of generated reports across GLIOMA, SOFT, UTERUS, with the same notation for error bars, bootstrap dots, and p-values.

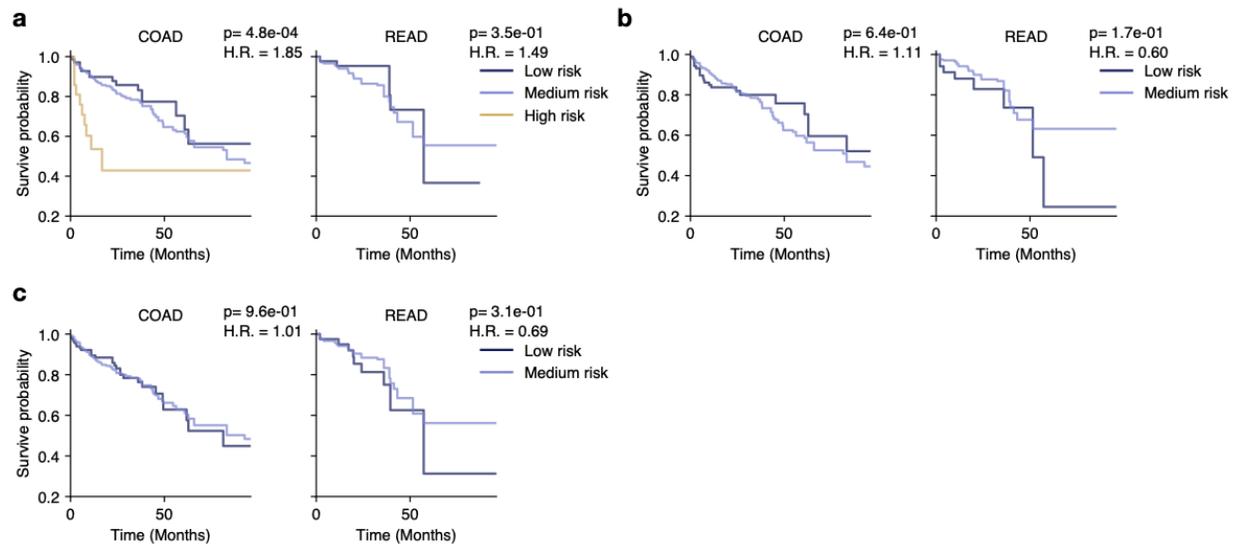

**Supplementary Figure 3. Supplementary results for report evaluation tasks. a-c,** COLON cohorts Kaplan–Meier curves comparing **a,** PathReasoning, **b,** single-turn GPT approach, and **c,** majority-vote approach. Curves are stratified into Low, Medium, and High risk.

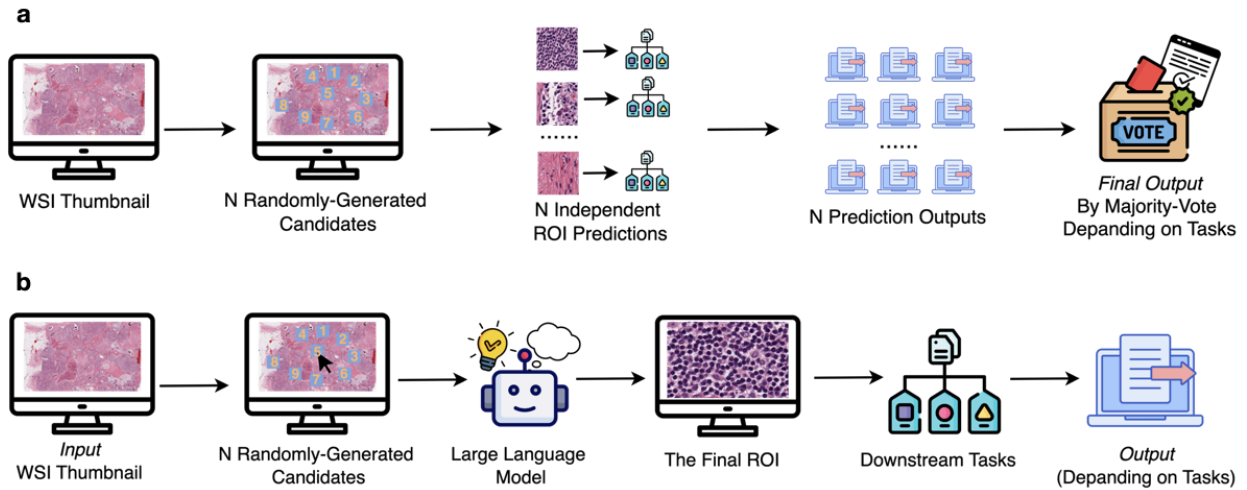

**Supplementary Figure 4. Schematic Illustration for baseline methods. a,** Schematic of the majority-vote approach, where N randomly generated ROIs from a WSI are each processed independently for downstream prediction tasks, and the final output is obtained by aggregating ROI-level results via majority vote. **b,** Schematic of the single-turn GPT approach, where an LLM selects a single ROI from N candidates on the WSI thumbnail, which is then used directly for downstream prediction tasks.

## STAR Methods

### EXPERIMENTAL MODEL AND SUBJECT DETAILS

**Whole-slide images (WSIs)**
We used diagnostic whole-slide images from The Cancer Genome Atlas (TCGA), covering 13 organ systems and 31 clinically relevant subtypes. All slides were scanned at 40× magnification and stored in SVS format. Subtype labels were derived from TCGA diagnostic metadata for each project (e.g., TCGA-BRCA, TCGA-LUAD). The 13 organ groups include BRCA, COLON, ESO, HEP, LUNG, RCC, GLIOMA, ADREN, CERVIX, PLEURA, SOFT, TESTIS, and UTERUS. Each group contains 2 to 4 clinically relevant subtypes, defined as follows:

**Breast (BRCA)**: Invasive Ductal Carcinoma (IDC) vs. Invasive Lobular Carcinoma (ILC)
**Colorectal (COLON)**: Colon Adenocarcinoma (COAD) vs. Rectal Adenocarcinoma (READ)
**Esophagogastric (ESO)**: Esophageal Adenocarcinoma (ESCA) vs. Esophageal Squamous Cell Carcinoma (ESCC) vs. Stomach Adenocarcinoma (STAD)
**Hepatobiliary (HEP)**: Cholangiocarcinoma (CHOL) vs. Hepatocellular Carcinoma (HCC)
**Lung (LUNG)**: Lung Adenocarcinoma (LUAD) vs. Lung Squamous Cell Carcinoma (LUSC)

**Kidney (RCC)**: Clear Cell RCC (CCRCC) vs. Chromophobe RCC (CHRCC) vs. Papillary RCC (PRCC)

**Glioma (GLIOMA)**: Glioblastoma Multiforme (GBM) vs. Oligodendroglioma (ODG)

**Adrenal (ADREN)**: Adrenocortical Carcinoma (ACC) vs. Pheochromocytoma (PHC)

**Cervix (CERVIX)**: Cervical Squamous Cell Carcinoma (CESC) vs. Endocervical Adenocarcinoma (ECAD)

**Pleura (PLEURA)**: Biphasic Mesothelioma (PLBMESO) vs. Epithelioid Mesothelioma (PLEMESO) vs. Sarcomatoid Mesothelioma (PLSMESO)

**Soft Tissue (SOFT)**: Dedifferentiated Liposarcoma (DDLS) vs. Leiomyosarcoma (LMS) vs. Myxofibrosarcoma (MFS) vs. Undifferentiated Pleomorphic Sarcoma (MFH)

**Testis (TESTIS)**: Seminoma (SEM) vs. Mixed Germ Cell Tumor (MGCT)

**Uterus (UTERUS)**: Uterine Endometrioid Carcinoma (UEC) vs. Uterine Serous Carcinoma (USC)

**Clinical reports**

To support text-based evaluation, we use the machine-readable TCGA pathology report corpus (9,523 reports, 32 tissues) produced via OCR, artifact removal, and normalization. This resource is publicly released by the Tatonetti Lab with accompanying paper/dataset describing curation and access (https://github.com/tatonetti-lab/tcga-path-reports).

**Checklists (structured clinical attributes)**

Cancer-specific enrollment / follow-up / sample submission forms from the TCGA Biospecimen Core Resource (BCR) at Nationwide Children's Biopathology Center provide structured attributes (e.g., architecture, mitoses, necrosis) that we use as ground truth for checklist evaluation. The BPC hosts the official TCGA forms catalog; example disease-specific enrollment forms are publicly posted.

**WSI visual question answering (WSI-VQA)**

We evaluate VQA using the WSI-VQA benchmark (ECCV 2024), which provides TCGA-grounded question–answer pairs (close-ended multiple-choice), dataset splits, and code; the repo also documents download and preprocessing of TCGA WSIs through the GDC (https://github.com/cpystan/WSI-VQA).

**Survival metadata.**

Overall Survival times and Overall Survival Status are derived from TCGA clinical data resources compiled for pan-cancer analyses; Overall Survival is used as a standard endpoint for risk stratification (high: < 12 mo, medium: 12–36 mo, low: > 36 mo) in our downstream survival tasks.

## METHOD DETAILS

**Details of PathReasoning Framework**

PathReasoning is a multi-stage ROI-selection agent that iteratively explores a WSI given a task prompt $q$ (e.g., subtyping, risk, reporting). At round $t$, the agent observes a state $s_t$ that aggregates a global thumbnail, previously selected ROIs, and the task prompt, and returns a next location and an optional stop decision. We structure this section around five design modules: candidate proposal, iterative exploration, early termination, and ROI extraction/memory update, and conclude with the output modes. (The mathematical policy notation used below is descriptive of a prompt-programmed GPT-4o[37], not the result of training a learned policy or optimizing a reward/value function.)

1. **Candidate proposal (rounds $t = 1 \sim 3$)**
   *Input:* Tissue foreground $\Omega_{\text{tissue}}$ and a fixed budget of $K = 20$ non-overlapping candidate coordinates $\mathcal{C} = \{c_i\}_{i=1}^{K} \subset [0,1]^2$ sampled from $\Omega_{\text{tissue}}$. To promote diversity, candidates satisfy a minimum spacing (0.01 in a 0-1 scale):
   $\mathcal{C} \subset \Omega_{\text{tissue}}, \quad \|c_i - c_j\|_2 \geq \delta \quad (\forall i \neq j).$
   *Process:* A GPT-4o–based policy $\pi_\theta$ scores each candidate under the current state $s_t$.
   *Output:* The selected coordinate and stop probability:
   $(a_t, \ p_{\text{stop}}^{(t)}) = \pi_\theta(s_t).$
   Candidate choice in the proposal rounds:
   $a_t = \arg\max_{c \in \mathcal{C}} f_\theta(s_t, c).$

2. **Iterative exploration (rounds $t \geq 4$)**
   *Input:* Updated state $s_t$ including all extracted ROIs up to round $t - 1$.
   *Process:* The policy operates directly in the continuous coordinate space without predefined proposals:
   $a_t \sim p_\theta(a \mid s_t), \ a_t \in [0,1]^2.$
   With greedy decoding this reduces to
   $a_t = \arg\max_{a \in [0,1]^2} f_\theta(s_t, a).$
   *Output:* The proposed coordinate $a_t$ and the associated $p_{\text{stop}}^{(t)}$.

3. **Early termination**
   The agent decides whether to stop after each round via a logistic gate:

$$p_{\text{stop}}^{(t)} = \sigma(g_\theta(s_t)).$$

Stop if:

$p_{\text{stop}}^{(t)} > \tau_{\text{stop}}$ or $t = T$.

We use a maximum number of rounds $T = 10$.

4. **ROI extraction and memory update**

    Given $a_t$, a high-resolution patch $\mathcal{R}_t$ is extracted at diagnostic magnification (level-0/1) and appended to the memory used to form $s_{t+1}$.

5. **Output modes**

    For classification-style tasks we use the final ROI $\mathcal{R}_T$ (or the last step before stopping). For generative or multi-turn reasoning tasks we retain a multi-ROI context $\{\mathcal{R}_1, \ldots, \mathcal{R}_k\}$ as evidence.

**Coordinate System and Agent I/O Structure**

We introduce the normalized coordinate system used to represent and navigate regions within WSIs. To enable spatial reasoning that is consistent across slides with varying shapes and resolutions, all coordinates in our system are normalized. Each Region of Interest (ROI) is defined by a center coordinate (x,y), where both values are scaled to lie in the range from 0 to 1, with (0,0) referring to the top-left corner of the slide and (1,1) referring to the bottom-right. This normalization allows the agent to operate independently of absolute pixel dimensions. For PathReasoning's I/O structure, the agent receives a prompt for each iteration, consisting of visual and textual inputs accumulated from earlier steps. The prompt includes:

1. A downsampled thumbnail image of the entire WSI at OpenSlide[44] level 3–4, (approximately 1.25x-2.5x magnification), with previously explored coordinates overlaid;
2. All previously selected zoomed-in ROI images extracted from OpenSlide level 0 (40x magnification) or level 1 (20x magnification), typically sized at 1024 × 1024 pixels;
3. A list of 20 new candidate coordinates (normalized to [0,1] range) in the first three rounds.

The agent is prompted to output its decision in the form of:

1. A coordinate (x,y) indicating the next region to examine (from a list of 20 candidates),
2. A short textual justification explaining its reasoning for choosing that location, and

3. An optional "TERMINATE" signal if it determines that sufficient information has been gathered.

Once the agent selects a coordinate, the system uses the OpenSlide library to extract a high-resolution region centered at that location. The ROI patch is then appended to the memory and visual prompt of the next iteration. This allows the agent to iteratively refine its spatial understanding by conditioning on both past choices and their corresponding visual content.

**Details of Task-Specific Implementations**

We evaluate PathReasoning across six distinct tasks: cancer subtyping, foundation model evaluation for cancer subtyping, diagnostic report generation, report checklist evaluation, WSI visual question answering (VQA), and survival risk prediction. All tasks operate directly on WSIs and use the selected ROIs to produce task-relevant outputs. Below we describe the methods we used for each task in detail.

1. **Cancer Subtyping**
   *Input:* A whole-slide image (WSI) with its organ group specified, and the organ-specific subtype set $\mathcal{Y}_g$. The agent supplies evidence ROIs (single ROI by default; multi-ROI optional). The task prompt explicitly enumerates $\mathcal{Y}_g$ and, for each subtype, provides concise histopathologic criteria (e.g., architectural pattern, cytologic features, keratinization or gland formation, stromal reaction, necrosis, mitotic activity) and instructs the model to return exactly one valid label from $\mathcal{Y}_g$.
   *Process:* The model consumes the ROI context and the structured prompt (label set + subtype-specific criteria) and deterministically selects one subtype. When multiple ROIs are used, the model will consider all the context of all ROIs to make a decision.
   *Output:* One subtype label per case drawn from $\mathcal{Y}_g$.
   *Evaluation Metrics:* The overall accuracy and macro F1 score for a cancer type is computed by:
   $$\text{Acc} = \frac{1}{N} \sum_{n=1}^{N} \mathbf{1}[\hat{y}_n = y_n]$$
   $$\text{F1}_{\text{macro}} = \frac{1}{C} \sum_{c=1}^{C} \frac{2 \, \text{Prec}_c \, \text{Rec}_c}{\text{Prec}_c + \text{Rec}_c}$$
   where $\mathcal{Y}_g$ represents the valid subtype set for organ group $g$; $\hat{y}_n$ represents the predicted label for case $n$; $y_n$ represents the ground-truth label; $N$ represents number of cases; $C$ represents the number of subtypes in that cancer type; and $\text{Prec}_c, \text{Rec}_c$ represent the precision/recall for class $c$ computed one-vs-rest.

2. **Foundation Model Evaluation For Cancer Subtyping**

   *Input:* For each slide in the subtyping task, PathReasoning returns one diagnostic ROI at level-0 (40×); we extract a $1024 \times 1024$ RGB patch centered at that coordinate.

   *Preprocessing:* The ROI patch is converted to the exact input format expected by the public pretrained tile encoder. We first resize the patch to 256 pixels on the short side with bicubic interpolation and then apply a centered crop to $224 \times 224$ using ImageNet[45] standard:
   $$I_{224} = \text{CenterCrop}_{224}\bigl(\text{Resize}_{256}(I_{\text{ROI}})\bigr).$$

   Then, the cropped image is converted to a floating-point tensor in [0,1] and normalized with ImageNet statistics:
   $$\tilde{I} = (I_{224} - \boldsymbol{\mu}) \oslash \boldsymbol{\sigma}, \quad \boldsymbol{\mu} = (0.485, 0.456, 0.406), \quad \boldsymbol{\sigma} = (0.229, 0.224, 0.225).$$

   Finally, the frozen pretrained tile encoder $\phi(\cdot)$ is used in evaluation mode to produce a single embedding per case:
   $$\mathbf{z} = \phi(\tilde{I}) \in \mathbb{R}^{1536}.$$

   *Process:* The encoder is treated strictly as a fixed feature extractor to test whether the single ROI chosen by the agent carries sufficient subtype signal. Two standard classifiers are evaluated on top of $\mathbf{z}$. A k-nearest neighbors classifier with cosine distance and $k = 10$ performs prediction by retrieving the ten nearest training embeddings and using their label distribution as class scores; this stage involves no parameter training beyond building the index. In parallel, a multinomial logistic regression head with L2 regularization (fixed $C = 1.0$) is trained on the training embeddings only (encoder frozen); at test time, the softmax outputs provide class scores. All steps run on CPU with fixed random seeds and the same patient-level splits as the main subtyping experiment.

   *Evaluation:* We report AUROC only, stratified per cancer type. For each organ group, one-vs-rest AUROC is computed for every subtype using the classifier's class scores (neighbor fractions for k-nearest neighbors and softmax probabilities for logistic regression) and summarized by the macro average across subtypes within that group. No additional post-processing is applied.

3. **Diagnostic Report Generation**

   *Input:* ROIs that are selected by PathReasoning with a specific report generation query , a reporting prompt that contains five few-shot examples drawn from real pathology reports in the TCGA dataset.

   *Process:* The language model generates a clinical-style narrative describing histological features and diagnostic impressions.

   *Output:* A professional clinical report for that patient by analyzing its pathology WSI.

*Evaluation Metrics:* We compare generated reports with their ground-truth counterparts using GPT as a judge. Mean model-judge score is computed on a 0–10 rubric:

$$S = \frac{1}{N} \sum_{n=1}^{N} s_n, \quad s_n \in [0, 10].$$

where $s_n$ is the GPT Eval Score for case n, $S$ is the mean score, and $N$ is the number of cases.

4. **Checklist Evaluation**

   *Input:* Generated report $\hat{R}$ and reference report $R^\star$; a cancer-specific checklist with $M$ discrete items; each item $m$ has a fixed option set $\mathcal{A}_m$ (e.g., {Yes, No} or multi-way choices).

   *Process:* For each checklist item, we run the same deterministic extraction instruction twice: once on the generated report to return a single option string, and once on the reference report to return its option. Both extractions are constrained to the item's predefined option list and require exact selection of one option. We then compare the two option strings to mark agreement for that item.

   *Output:* The method returns a predicted option vector and its reference counterpart, along with an optional per-item agreement vector:
   $\hat{\mathbf{z}} = (\hat{z}_1, \ldots, \hat{z}_M), \qquad \mathbf{z}^\star = (z_1^\star, \ldots, z_M^\star).$
   We also expose the agreement vector (1 = match, 0 = mismatch) per case:
   $\mathbf{a} = (\text{agree}_1, \ldots, \text{agree}_M), \qquad \text{agree}_m = \mathbf{1}[\hat{z}_m = z_m^\star].$

   *Evaluation Metrics:* For each case $n$ and checklist item $m$, we extract the model's agree vector above and calculate the single case's accuracy on a checklist is the mean agreement across all items in that checklist by using:

   $$\text{Acc}_{\text{case},n} = \frac{1}{M_n} \sum_{m=1}^{M_n} \text{agree}_{n,m}$$

   Then, we can calculate the dataset's overall accuracy averages case-level accuracies across all $N$ cases by using:

   $$\text{Acc} = \frac{1}{N} \sum_{n=1}^{N} \text{Acc}_{\text{case},n}$$

   This two-stage averaging ensures that each case contributes equally, regardless of how many checklist items it contains.

5. **WSI Visual Question Answering (VQA):**

   *Input:* A natural-language question $q$ with a finite multiple-choice answer set $\{a_1, \ldots, a_J\}$, and evidence ROIs $\{\mathcal{R}_{1:k}\}$ selected by the agent. For analysis, each question is tagged into coarse clinical categories (e.g., diagnosis, staging, grading/proliferation, structure, margins, biomarkers, other) using keyword-based rules on the question text; the keyword lists are predefined and fixed.

*Process:* The model reads $q$ and the ROI context then performs constrained multiple-choice selection: it must output exactly one option from the provided answer set.

*Output:* One selected answer for the question, denoted $\hat{a}$. For per-category analysis, predictions are grouped by the keyword-derived tags of their questions.

*Evaluation Metrics:* Overall accuracy on the evaluation set:

$$\text{Acc} = \frac{1}{N} \sum_{n=1}^{N} \mathbf{1}[\hat{a}_n = a_n^\star]$$

where $\hat{a}_n$ represents the predicted answer for question $n$; $a_n^\star$ represents the ground-truth answer; and $N$ represents the number of questions. Per-category accuracies are computed by filtering the set of questions with the corresponding keyword tag and applying the same accuracy formula within each subset.

6. **Survival Risk Prediction**

    *Input:* Evidence ROIs $\{\mathcal{R}_{1:k}\}$ selected by the agent; A few-shot example set $\mathcal{E}_{\text{risk}}$ that contains three pairs of selected ROI and its risk level; ordinal risk labels $\{0, 1, 2\}$ corresponding to low, medium, and high risk. Survival time and censoring indicators are used only for evaluation.

    *Process:* The model consumes the ROI context and the exemplar set and deterministically assigns one ordinal risk level to each case.

    *Output:* A single predicted risk level per case: $\hat{y} \in \{0, 1, 2\}$.

    *Evaluation Metrics:* We evaluate the overall log-rank p-value across the $G$ predicted groups in the Kaplan–Meier stratification computed on the predicted groups:

    $$(\text{KM log-rank}) \quad \chi^2 = (\mathbf{O} - \mathbf{E})^\top \mathbf{V}^{-1} (\mathbf{O} - \mathbf{E}), \qquad p = 1 - F_{\chi^2_{G-1}}(\chi^2).$$

    where $\mathbf{O}, \mathbf{E}$ are the observed and expected event-count vectors aggregated over event times under the null of equal hazards; $\mathbf{V}$ is the Greenwood variance–covariance matrix for the log-rank test; $F_{\chi^2_{G-1}}$ is the chi-squared CDF with $G - 1$ degrees of freedom.

**QUANTIFICATION AND STATISTICAL ANALYSIS**

**Comparison approaches downstream tasks**

For Random tiles methods with majority aggregation (majority-vote approach), we construct a non-iterative reference that omits spatial reasoning for each WSI. We uniformly sample a fixed set of $K = 21$ non-overlapping, non-blank foreground coordinates at level-0 (40×) and extract $1024 \times 1024$ RGB patches centered at those locations. Each patch is processed independently by the same downstream head used in the corresponding task, with identical preprocessing, prompts, and patient-level splits as the main method. No iterative refinement, memory accumulation, or coordinate proposal is performed beyond the initial random sampling. Aggregation rules are task-specific but deterministic. For classification task, such as cancer subtyping, WSI-VQA, and survival risk prediction, we aggregate the $K$ independent predictions by majority vote:

$$\hat{y} = \arg\max_{c \in \mathcal{Y}} \sum_{i=1}^{K} \mathbf{1}\big[f(I_i) = c\big].$$

where $I_i$ denotes the $i$-th sampled ROI patch from the WSI; $f(\cdot)$ is the task-specific predictor that maps a patch to a discrete label in $\mathcal{Y}$; $\mathcal{Y}$ is the set of admissible classes (e.g., subtypes, risk levels, or VQA answers); $\mathbf{1}[\cdot]$ is the indicator function that outputs either $1$ or $0$; and $\hat{y}$ is the final slide-level prediction obtained by majority vote (ties broken by a fixed class order). For diagnostic report generation task, we produce one report per patch and average the GPT-eval scores across the $K$ reports to obtain the case-level score:

$$S_{\text{case}} = \frac{1}{K} \sum_{i=1}^{K} s_i.$$

where $s_i$ is the GPT-eval score assigned to the report generated from the $i$-th sampled ROI/patch. For checklist evaluation we compute item-wise agreement for each report and first form the per-report case accuracy, then average across the $K$ reports for the slide:

$$\text{Acc}_{\text{case}}^{(i)} = \frac{1}{M} \sum_{m=1}^{M} \mathbf{1}\big[\hat{z}_m^{(i)} = z_m^\star\big], \qquad \text{Acc}_{\text{case}} = \frac{1}{K} \sum_{i=1}^{K} \text{Acc}_{\text{case}}^{(i)}.$$

where $M$ is the number of checklist items for the cancer type; $\hat{z}_m^{(i)}$ is the option predicted for item $m$ from the $i$-th generated report; $z_m^\star$ is the ground-truth option for item $m$;.

On the other hand, for the GPT one-shot coordinate selection (single-turn GPT approach), we preserve a single discrete selection from proposals while removing multi-round exploration. For each slide we draw $K = 20$ normalized candidate coordinates from tissue foreground (non-overlapping, non-blank) and present the thumbnail plus the candidate list in the task prompt. The language model selects exactly one coordinate; we extract a single $1024 \times 1024$ level-0 patch at that location and run the downstream head once to produce the task output. All other settings, including ROI size, preprocessing, prompts, and patient-level splits, still match the main PathReasoning method for a controlled comparison.

**Evaluation Settings and Statistical Analysis**

To ensure a like-for-like comparison, when PathReasoning is configured to return $m$ ROIs per slide, all baselines are required to produce exactly $m$ ROIs as well. For the majority-vote approach, we uniformly sample $m$ non-overlapping, non-blank foreground locations (subject to the same spacing $\delta$) and process the mmm patches independently. For the single-turn GPT approach, we select $m$ coordinates without replacement from the candidate list by iterating the one-shot selection $m$ times while not providing previously selected patches as additional context, thereby avoiding iterative memory. Downstream aggregation strictly matches the task rule used by PathReasoning (e.g., majority vote for classification/VQA, mean of judge scores for report generation, two-stage averaging for checklist). Unless otherwise specified, all methods share the same patient-level splits, preprocessing, prompts, and downstream heads; only the ROI-selection policy differs. We report per-task metrics as defined in the Task-Specific Implementations. Across tasks, uncertainty is quantified by case-level bootstrap[46] ($B = 1000$); we report mean and 95% CIs. Paired significance tests are used when comparing PathReasoning to the strongest baseline on the same cases.

**Resource Availability**

**1. Lead Contact**

The first author Kunpeng Zhang can be reached via email: kzhang27@cs.washington.edu. The corresponding author Sheng Wang can be reached via email: swang@cs.washington.edu

**2. Materials Availability**

We have provided the relevant materials in the Data Availability.

**3. Data and Code Availability**

PathReasoning code is available at https://github.com/Erickunpeng/PathReasoning.

# Appendix: Evaluation Prompt Specification

### 1: Prompt for PathReasoing's System Message

```
Find a 1024px x 1024px region of interest (ROI) on the whole slide image
(WSI) based on the user's query. Select the most relevant ROI by defining its
bounding box and downsample level. The bounding box is determined by its
top-left corner (x, y) relative to the top-left corner of the WSI. For
example, x=0.5, y=0.5 represents the center of the WSI.
Adjust the downsample level to zoom in or out:
- Zoom in for more detail with a lower level (e.g., level=0 for highest
magnification).
- Zoom out for a larger area with a higher level (e.g., level=1 or above).
The maximum downsample level of the WSI will be provided. An overview of the
WSI and the current ROI, highlighted by a bounding box, will also be shown.
Assess if the current ROI meets the user's needs. If it does, respond with
"TERMINATE." If not, suggest a new ROI in the format: <<x, y, level>>.
To check different areas, adjust the coordinates. For example:
- To check the left area from the current location (x=0.5, y=0.5), use
(x=0.4, y=0.5).
- To check the lower area, use (x=0.5, y=0.6).
Ensure to check multiple areas in the slide to find the best region of
interest.
In each response, provide a brief medical reasoning (one sentence) explaining
why the selected region is appropriate. Describe any notable cellular or
structural features that support your decision.
End your response using the following two-line format:
<<x=…, y=…, level=…>>
For example:
"This region displays dense nuclear atypia and irregular gland formation,
consistent with carcinoma."
<<x=0.43, y=0.62, level=0>>
Make sure the coordinate format exactly matches <<x=…, y=…, level=…>> for
automatic parsing.
```

### 2: Prompt for Iteration Messages

Cancer subtyping (BRCA for example):

```
"What is the cancer subtype of this slide? Is it Invasive Ductal Carcinoma
(IDC) or Invasive Lobular Carcinoma (ILC)?"
```

Clinical report generation:

```
Generate a concise pathology report for this {cancer_type} cancer slide.
```

```
Select an ROI that best captures diagnostic features such as tumor
architecture, cellular morphology, and relevant markers.
```

Survival Risk Prediction:

```
What is the survival risk level for this {cancer_type} cancer image?
Select from the following options: Low (>36 months), Intermediate (12-36
months), or High (<12 months).
```

### 3: Prompt for Cancer Subtyping

Subtyping prediction:

```
You are provided {num_images} histological slides from a tissue biopsy. Each
slide highlights a specific region of interest (ROI) containing key cellular
structures. Your task is to classify the slides into one of the following
subtypes: [SUBTYPE NAMES].
Use the detailed histological features below to guide your decision-making:
[SUBTYPE DESCRIPTIONS HERE]
Analyze all {num_images} slides provided and determine the most likely
subtype based on the consistency of features observed across the regions.
Considering these characteristics, **provide only the classification word** –
[e.g., 'IDC' or 'ILC'] – without any additional text or explanation. You have
to make a decision even though you are unsure.
```

Subtypes Descriptions:

```
BRCA (Breast Cancer)
IDC: Typically presents with glandular or duct-like structures, often forming
cohesive nests or sheets of cells. Central necrosis may be evident in ducts.
Cells exhibit nuclear pleomorphism, prominent nucleoli, and frequent mitotic
figures. Tumor cells are typically arranged in dense clusters, forming clear
ductal structures.
ILC: Displays single-file infiltration patterns of tumor cells into the
stroma. Cells may be arranged in targetoid patterns around ducts and lack
duct formation. Reduced or absent E-cadherin expression is a key feature.
Tumor cells are generally small with uniform nuclei and inconspicuous
nucleoli, showing a diffuse and ill-defined growth pattern.

LUNG (Lung Cancer)
LUAD: Typically shows glandular differentiation with acinar, papillary, or
lepidic growth patterns. Cells may contain intracellular mucin, and mucin
production is often evident. Tumor cells have round nuclei with prominent
nucleoli. Commonly arises in the peripheral regions of the lung and may be
```

associated with scar tissue. Tumor cells are arranged in loose clusters forming glandular or papillary structures.
LUSC: Exhibits squamous differentiation, including keratin pearls, intercellular bridges, and polygonal cells with abundant eosinophilic cytoplasm. Nuclear pleomorphism and hyperchromasia are common. Tumor cells are arranged in cohesive nests and sheets, often originating in central lung regions and associated with bronchial structures.

COLON (Colorectal Cancer)
COAD: Frequently arises in the proximal or distal colon. Tumors exhibit glandular or tubular structures with varying degrees of differentiation. Mucin pools and irregular gland formation are common. Cells exhibit moderate-to-high nuclear atypia, and invasion into the muscularis propria and pericolonic fat is often observed. Tumor cells form loose clusters within the glandular or tubular architecture.
READ: Typically originates in the rectum and shares features with COAD, including irregular glands and mucin production. However, READ tends to have circumferential involvement and may more frequently exhibit perineural invasion. Tumor deposits in the mesorectal fat are more commonly observed. Tumor cells are arranged in cohesive clusters forming glandular or tubular structures.

RCC (Renal Cell Carcinoma)
CCRCC: Characterized by clear cytoplasm due to glycogen and lipid accumulation. Cells are arranged in nests or alveolar structures surrounded by a delicate vascular network. The tumor often exhibits a golden-yellow appearance grossly due to lipid content. Nuclei show varying degrees of atypia, with prominent nucleoli in higher grades. Hemorrhage, necrosis, and cystic degeneration are common histological findings.
PRCC: Displays papillary or tubular-papillary structures with fibrovascular cores. The tumor cells are cuboidal to columnar and often contain foamy macrophages within the fibrovascular cores. Basophilic, eosinophilic, or oncocytic cytoplasm may be present. Psammoma bodies are frequently observed. The stroma may show edema, and hemorrhage is not uncommon. Multinucleated giant cells may occasionally be identified.
CHRCC: Composed of large polygonal cells with pale, eosinophilic cytoplasm and distinct cell borders. Tumor cells often demonstrate a perinuclear halo and finely reticulated cytoplasm resembling plant cells. The nuclei are irregular, with perinuclear clearing. The tumor architecture is predominantly solid, with sheets or trabeculae, and may include small, focally arranged tubules. Cytoplasmic granularity is another distinguishing feature.

ESO (Esophageal/Gastric Cancer)

ESCA: Typically arises in the lower esophagus and is associated with Barrett's esophagus. Tumors show glandular differentiation, often forming tubular or papillary structures. Cells may exhibit mucin production and have prominent nucleoli.

ESCC: Occurs in the middle or upper esophagus. Displays squamous differentiation with features like keratin pearls, intercellular bridges, and polygonal cells. Nuclei are hyperchromatic with irregular contours.

STAD: Frequently exhibits glandular or tubular structures, with varying degrees of differentiation. Tumors may show mucin production, nuclear pleomorphism, and invasion into the gastric wall. Tumor cells are typically arranged in irregular glands or sheets.

HEP (Liver Cancer)
CHOL: Arises from the bile ducts and typically forms glandular or tubular structures with dense fibrous stroma. Tumor cells are cuboidal to columnar with prominent nucleoli and may show mucin production. Desmoplastic stromal reaction is a hallmark feature.

HCC: Originates in hepatocytes, with tumor cells arranged in trabecular, pseudoacinar, or solid patterns. Cells often have abundant eosinophilic cytoplasm, round nuclei, and prominent nucleoli. Intracytoplasmic bile or Mallory bodies may be present. Endothelial-lined vascular spaces are frequently observed.

GLIOMA (Brain Tumors)
GBM: High-grade astrocytic tumor characterized by marked cellular atypia, high mitotic activity, necrosis, and microvascular proliferation. Necrotic areas are often surrounded by hypercellular regions (pseudopalisading necrosis). Cells may appear pleomorphic with hyperchromatic nuclei.

ODG: Low- to intermediate-grade glioma with tumor cells forming a 'fried-egg' appearance due to clear cytoplasmic halos. Tumors often contain delicate branching capillaries resembling a 'chicken-wire' pattern. Calcifications and uniform, round nuclei are commonly observed.

ADREN (Adrenal Tumors)
ACC: Malignant tumor of the adrenal cortex, often showing high nuclear pleomorphism, atypical mitoses, necrosis, and increased mitotic rate. Cells may have eosinophilic or clear cytoplasm, arranged in sheets or nests.

PHC: Tumor arising from chromaffin cells of the adrenal medulla, characterized by large polygonal cells with abundant granular cytoplasm. Zellballen pattern with highly vascularized stroma is common, and nuclei are round with a salt-and-pepper chromatin pattern.

CERVIX (Cervical Cancer)

CESC: Derived from the squamous epithelium, showing nests of malignant cells with keratinization and intercellular bridges. Tumor cells may have pleomorphic, hyperchromatic nuclei and eosinophilic cytoplasm.
ECAD: Glandular malignancy arising from the endocervical canal, characterized by columnar cells forming irregular glands with mucin production. Nuclear atypia and mitotic figures are commonly seen.

PLEURA (Mesothelioma)
PLBMESO: Contains both epithelioid and sarcomatoid components, with at least 10% of each pattern present. Epithelioid areas appear glandular or tubulopapillary, while sarcomatoid regions consist of spindle cells.
PLEMESO: Composed of uniform, polygonal tumor cells forming tubulopapillary, trabecular, or solid patterns. Cells have round nuclei, prominent nucleoli, and abundant eosinophilic cytoplasm.
PLSMESO: Highly aggressive subtype with elongated spindle cells arranged in fascicles. Lacks glandular differentiation and often resembles fibrosarcoma or other soft tissue sarcomas.

SOFT (Soft Tissue Sarcoma)
DDLS: Displays well-differentiated liposarcoma adjacent to high-grade non-lipogenic sarcoma. The dedifferentiated areas contain pleomorphic spindle cells with high mitotic activity.
LMS: Malignant smooth muscle tumor composed of elongated spindle cells with cigar-shaped nuclei. Tumors exhibit eosinophilic cytoplasm, nuclear pleomorphism, and frequent mitoses.
MFS: Characterized by myxoid stroma with curvilinear vasculature and scattered pleomorphic tumor cells. Low-grade tumors have more myxoid matrix, while high-grade variants show increased cellular atypia.
MFH: Poorly differentiated sarcoma with bizarre, pleomorphic cells arranged in storiform or haphazard patterns. Features include high mitotic activity and tumor giant cells.

TESTIS (Testicular Tumors):
SEM: Composed of uniform polygonal tumor cells with clear cytoplasm and centrally located nuclei. Tumors have fibrous septa infiltrated by lymphocytes and may contain granulomas.
MGCT: Contains multiple germ cell tumor components, such as embryonal carcinoma, yolk sac tumor, choriocarcinoma, and teratoma. Each component has distinct histological features, including glandular, solid, or papillary growth patterns.

UTERUS (Endometrial Cancer):

```
UEC: Common endometrial malignancy with glandular structures resembling
normal endometrium. Tumor cells have round to oval nuclei with variable
atypia and occasional squamous differentiation.
USC: Aggressive high-grade carcinoma with papillary or glandular
architecture. Tumor cells exhibit marked pleomorphism, high mitotic index,
and prominent nucleoli.
```

**4: Prompt for Visual Question Answering**

```
"You are a medical AI assistant trained to analyze pathology slides and
answer multiple-choice questions related to cancer diagnosis. You are
provided {num_images} histological slides from a tissue biopsy. Each slide
highlights a specific region of interest (ROI) containing key cellular
structures. For each question, select the most appropriate answer from the
given choices. If you are uncertain, select the answer that is the closest
match based on available information. Your response must strictly follow the
order of the questions, and answers should be separated by a comma. Do not
provide any explanations or additional information.
{questions_block}
Analyze all {num_images} slides provided and determine the most likely answer
for each question.
Answers:"
```

**5: Prompt for Clinical Report Generation**

Generating clinical reports:

```
"Generate a comprehensive pathology report for a patient diagnosed with
{cancer_type}. This report should be professional, medically accurate, and
well-structured. You are provided with some real-world pathology reports as
reference. These examples are meant to offer insights into the type of
information that might be included in such reports, but you do NOT need to
follow their format or structure exactly. Instead, generate a report using
your own medical knowledge, ensuring it is detailed, logical, and clinically
relevant. Here are some pathology report examples for reference: [EXAMPLES].
Now, based on your medical expertise, generate a detailed pathology report
for a patient with the same cancer type. Make sure your report is structured
professionally, with accurate clinical descriptions and findings."
```

GPT-eval Score:

```
"You are an expert in scientific pathology report evaluation. Your task is to
compare two pathology reports and assign a similarity score on a scale from 0
to 10. A score of 10 indicates that the reports describe nearly identical
medical findings, whereas a score of 0 means they discuss completely
```

different content. Ignore irrelevant details such as patient name, sample ID, date, physician name, and other administrative information. Although the reports may differ in formatting, focus only on comparing their medical content, including diagnoses, observations, and clinical details. Provide only a single numerical score from 0 to 10, without explanation."

Checklist evaluation:

"You are an expert pathologist simulating the process of filling out a TCGA enrollment form based on pathology reports. You will compare the reference pathology report and the candidate pathology report to evaluate their consistency. Each question corresponds to a section in the form, and the choices are the available options. Your task is to determine whether the candidate report provides the same answer as the reference report for each question.
Reference Report:
"""{reference_text}"""
Candidate Report:
"""{candidate_text}"""
Checklist Questions:
{formatted_questions}
Instructions:
- For each question, determine if the candidate report provides the same answer as the reference report.
- If the answers are identical, return 0. If they are different, return 1.
- Ignore minor wording differences, focus on the meaning.
- If the candidate report does not provide an answer, assume it differs from the reference and mark it as 1.
- Only return a JSON array of 0s and 1s, strictly in order, without any additional text.
- Do NOT include any extra text. The output must ONLY be a valid JSON array.
Output Format (Example):
[0, 0, 1, 1, 0, …]
Now, generate your response."

### 6. Prompt for Survival Risk Prediction

"You are an expert pathologist specializing in histological image analysis. Your task is to predict the risk level (0, 1, or 2) for a patient based on several provided regions of interest (ROI) from a histopathology slide.
The risk levels are defined as follows:
- 0 = Low risk (long survival time, e.g., > 36 months)
- 1 = Medium risk (moderate survival time, e.g., 12–36 months)
- 2 = High risk (short survival time, e.g., < 12 months)

```
Instruction:
1. The first 3 images are example ROI images with known risk levels.
2. The remaining images are new ROI images from a patient with {cancer_type}
cancer. Your task is to analyze all remaining ROIs and assign a risk level
(0, 1, or 2) for this slide.
3. Return only a SINGLE predicted risk level for all remaining images (except
the first three images). For example: 1
4. ONLY return a SINGLE number from 0, 1, and 2, even though you are not
sure. Do NOT include ANY other information in the response!
Few-shot Example Risk Levels:
Example 1: Risk Level = {…}
Example 2: Risk Level = {…}
Example 3: Risk Level = {…}
Now analyze the remaining images (Starts at the fourth image) and return only
the predicted risk level."
```